\documentclass{article}

\usepackage[final]{arxiv}

\def\mB{{\bm{B}}}

\DeclareMathAlphabet{\mathsfit}{\encodingdefault}{\sfdefault}{m}{sl}
\SetMathAlphabet{\mathsfit}{bold}{\encodingdefault}{\sfdefault}{bx}{n}

\def\gN{{\mathcal{N}}}

\def\sR{{\mathbb{R}}}

\newcommand{\sigmoid}{\Phi}

\usepackage{amsfonts}
\usepackage{amsmath,bm}

\newtheorem{theorem}{Theorem}

\newtheorem{corollary}{Corollary}
\newtheorem{definition}{Definition}

\newtheorem{remark}{Remark}

\newcommand{\expectation}[2]{\underset{#1}{\mathbb{E}}\left[#2\right]}
\newcommand{\norm}[1]{\left\lVert#1\right\rVert}

\usepackage{xargs}                      
\usepackage[colorinlistoftodos,prependcaption]{todonotes}
\newcommandx{\red}[2][1=]{\todo[linecolor=red,backgroundcolor=red!25,bordercolor=red,#1]{#2}}
\newcommandx{\blue}[2][1=]{\todo[linecolor=blue,backgroundcolor=blue!25,bordercolor=blue,#1]{#2}}
\newcommandx{\green}[2][1=]{\todo[linecolor=OliveGreen,backgroundcolor=OliveGreen!25,bordercolor=OliveGreen,#1]{#2}}
\newcommandx{\purple}[2][1=]{\todo[linecolor=Plum,backgroundcolor=Plum!25,bordercolor=Plum,#1]{#2}}

\usepackage{mathrsfs}

\newcommand{\ppa}{\partial}
\newcommand{\der}[2]{\frac{\ppa #1}{\ppa #2}}
\newcommandx{\del}[1]{\dfrac{\mathrm{d} }{\mathrm{d} #1}}

\newcommand{\lp}{\left(}
\newcommand{\rp}{\right)}
\newcommand{\lb}{\left[}
\newcommand{\rb}{\right]}

\newcommand{\linnerprod}{\left\langle}
\newcommand{\rinnerprod}{\right\rangle}

\newcommand{\trace}{\mathrm{tr}}

\newcommand*{\comb}[2]{{}^{#1}C_{#2}}

\usepackage[utf8]{inputenc} 
\usepackage[T1]{fontenc}    
\usepackage{hyperref}       
\usepackage{url}            
\usepackage{booktabs}       
\usepackage{amsfonts}       
\usepackage{nicefrac}       
\usepackage{microtype}      
\usepackage{xcolor}         
\usepackage{natbib}
\usepackage{stackrel}
\usepackage{comment}
\usepackage{makecell}

\title{New Insights into Graph Convolutional Networks 
\\using Neural Tangent Kernels}

\author{Mahalakshmi Sabanayagam\\
Technical University of Munich\\
\And
Pascal Esser \\
Technical University of Munich
\AND
Debarghya Ghoshdastidar \\
Technical University of Munich
}

\begin{document}

\maketitle

\begin{abstract}

Graph Convolutional Networks (GCNs) have emerged as powerful tools for learning on network structured data.
Although empirically successful, GCNs exhibit certain behaviour that has no rigorous explanation---for instance, the performance of GCNs significantly degrades with increasing network depth, whereas it improves marginally with depth using skip connections.

This paper focuses on semi-supervised learning on graphs, and explains the above observations through the lens of Neural Tangent Kernels (NTKs).
%
%
We derive NTKs corresponding to infinitely wide GCNs (with and without skip connections). 
Subsequently, we use the derived NTKs to identify that, with suitable normalisation, network depth does not always drastically reduce the performance of GCNs---a fact that we also validate through extensive simulation.
Furthermore, we propose NTK as an efficient `surrogate model' for GCNs that does not suffer from performance fluctuations due to hyper-parameter tuning since it is a hyper-parameter free deterministic kernel.
The efficacy of this idea is demonstrated through a comparison of different skip connections for GCNs using the surrogate NTKs.
\end{abstract}

\section{Introduction}
\label{sec:intro}
Graph structured data are ubiquitous in various domains, including social network analysis, bioinformatics, communications engineering among others.
In recent years, graph neural networks 
have become an indisputable choice for various learning problems on graphs, and have been employed in a wide range of applications across domains.
Several variants of graph neural networks have been proposed, including graph convolutional network \citep{kipf2017semi}, graph recurrent network \citep{scarselli2008graph,li2015gated}, graph attention network \citep{velivckovic2017graph}, to name a few.
%
The popularity of graph neural networks can be attributed to their ability to tackle two conceptually different learning problems on graphs.
In \emph{supervised learning on graphs}, each data instance is a graph and the goal is to predict a label for each graph (for example, a protein structure). In contrast, \emph{semi-supervised learning on graphs} (also called \emph{node classification} or \emph{graph transduction}) refers to the problem of predicting the labels of nodes in a single graph. 
For instance, given the memberships of a few individuals in a social network, the goal is to predict affiliations of others.

This work focuses on the latter problem of semi-supervised learning. 
GCNs, along with its variants that locally aggregate information in the neighbourhood of each node, have proved to be superior methods in practice \citep{defferrard2016convolutional,kipf2017semi,chen2018fastgcn,wu2019simplifying,chenWHDL2020gcnii}, outperforming classical, and well-studied, graph embedding based approaches.
Among the different variants of GCNs, we focus on the methods based on approximations of spectral graph convolutions \citep{defferrard2016convolutional,kipf2017semi}, rather than spatial graph convolutions \citep{hamilton2017inductive,xu2018powerful}.
Surprisingly, these papers suggest shallow networks for the best performance, and unlike the standard neural networks that gain advantage with depth, the performance of GCN has been reported to decrease for deeper nets. This appears to be due to the over smoothing effect of applying many convolutions, that is, with repeated application of the graph diffusion operator in each layer, the feature information gets averaged out to a degree where it becomes uninformative.
As a solution to this, \citet{chenWHDL2020gcnii} and \citet{kipf2017semi} proposed different formulations of skip connections in GCNs that overcome the smoothing effect and thus outperform the vanilla GCN empirically.
These networks achieve state-of-the-art results by directly operating on graphs which enables effective capturing of the complex structural information as well as the features associated with the entities.
However, similar to standard neural networks, tuning the hyper-parameters is particularly hard due to the highly non-convex objective function and the over-parameterised setup making it computationally intense. 
As a result, there is no theoretical framework that supports rigorous analysis of graph neural networks.
Furthermore, the graph convolutions increase the difficulty of analysis.
Motivated by this, we are interested in a more formal approach to analyze GCNs and, specifically, to understand the influence of depth.

Explaining the empirical evidence of deep neural networks through mathematical rigour is an active area of research. In contrast, theoretical analysis of graph neural networks has been limited in the literature. 
From the perspective of learning theory, generalisation error bounds have been derived for graph neural networks using complexity measures like VC Dimension  and Rademacher complexity \citep{Scarselli_2018_Neuralnet,Garg_2020_arxiv}.
However, it is often debated whether generalisation error bounds can explain the performance of deep neural networks \citep{NIPS2017_Neyshabur}.
Another line of research relies on the  connection between graph convolutions and belief propagation \citep{Dai_2016_ICML} to analyse the behaviour of graph neural networks in both supervised and semi-supervised settings using cavity methods and mean field approaches \citep{Zhou_2020_PhysRevResearch,Kawamoto_2019_StatisticalMechanics,  Chen_2019_ICLR}. However, the above lines of research do not completely explain the empirical trends observed in GCNs, especially with regards to the aspects analysed in our work.

In this paper, we explain the empirically observed trends of GCNs using the recently introduced \emph{Neural Tangent Kernel} (NTK) \citep{jacot2018neural}.
NTK was proposed to describe the behaviour and generalisation properties of randomly initialised fully connected neural networks during training by gradient descent with infinitesimally small learning rate.
\citet{jacot2018neural} also showed that, as the network width increases, the change in the kernel during training decreases and hence, asymptotically, one may replace an infinitely wide neural network by a deterministic kernel machine, where the kernel (NTK) is defined by the gradient of the network with respect to its parameters as 
\begin{align}
    \Theta(x,x^\prime) = \expectation{W \sim \mathcal{N}}{\linnerprod \der{F(W,x)}{W} , \der{F(W,x^\prime)}{W} \rinnerprod}. \label{eq:ntk}
\end{align}
Here $F(W,x)$ represents the output of the network at data point $x$ and the expectation is with respect to $W$, that is, all the parameters of the network randomly sampled from Gaussian distribution $\gN$.
There has been criticism of the `infinite width' assumption being too strong to model real (finite width) neural networks, and empirical results show that NTK often performs worse than the practical networks \citep{arora2019exact,lee2019wide}.
Nevertheless, theoretical insights on neural network training gained from NTK have proved to be valuable, particularly in showing how gradient descent can achieve good generalisation properties \citep{du2019gradient}.
%
%
%
Subsequent works have derived NTK to analyse different neural network architectures in infinite width limit, including convolutional networks, recurrent networks among others \citep{arora2019exact,du2018gradient,du2019gradient,alemohammad2020recurrent}.
%
The most relevant work in the context of our discussion is the work of \citet{du2019graph} that derived NTK for graph neural networks in the supervised setting (each graph is a data instance to be classified) and empirically showed that graph NTK outperforms most graph neural networks as well as other graph kernels for the problem of graph classification. 

\textbf{Focus of this paper and contributions.}
The focus of the present paper differs from existing work on graph NTK \citep{du2019graph} in two key aspects---we derive NTK for semi-supervised node classification and, more importantly,
we use the derived NTKs to rigorously analyse corresponding GCN architectures and demonstrate the cause for surprising trends observed empirically in GCNs, as opposed to standard deep neural networks.
More precisely, we make the following contributions:

%
{\bf1.} In Section~\ref{sec:vanilla_gcn}, we derive the NTKs for GCNs used in semi-supervised node classification \citep{kipf2017semi,wu2019simplifying} 
in infinite width limit. 
In contrast to simplifying assumptions in most NTKs derivations, we allow a non-linear (sigmoid) pooling in the last layer---a natural choice in practical networks for binary classification.
Using the derived NTK and through extensive simulation, we show that the performance of GCN varies considerably for different hyper-parameters, but NTK captures the general trend of the best possible performance of GCN.
%

{\bf2.} Due to the observation that NTK is a hyper-parameter free alternative to GCN that approximates the behaviour of GCNs, we suggest NTK as an efficient surrogate for GCN that could be used to identify the optimal network architecture.
We demonstrate this idea in Section~\ref{sec:skip_gcn} by deriving the NTKs corresponding to GCNs with different skip connections \citep{chenWHDL2020gcnii,kipf2017semi}, and we make recommendation on the skip connection for improved performance through empirical studies of the NTKs. 
The  NTK surrogate can be further used to assess the relative importance of structure and feature information in a graph dataset.

{\bf3.} In Section~\ref{sec:norm_role}, we use our NTK based analysis to investigate the popular belief that the performance of vanilla GCN degrades drastically with increasing network depth. 
We demonstrate that this observation is due to instabilities in the network training, which results in performance fluctuations of vanilla GCN,
and that can be addressed by appropriate normalisation of the features at each level.
The fluctuations can also be reduced by adding skip connections, even without appropriate normalisation.

{\bf4.} In Section~\ref{sec:convergence}, we explain an empirical finding---unlike vanilla GCNs, the performance of NTK for certain skip connections converge with network depth. This is because the NTKs for skip connections converge with network depth, whereas this is less prominent in the case of NTK for vanilla GCNs.


We conclude in Section~\ref{sec:conclusion}, and provide the NTK derivations and further experimental details in the appendix.

\textbf{Notation.}
We represent the matrix Hadamard (entry-wise) product by $\odot$ and the scalar product by $\linnerprod.,.\rinnerprod$. 
We use $M^{\odot k}$ to denote Hadamard product of matrix $M$ with itself repeated $k$ times.
Let $\gN(\mu,\Sigma)$ be Gaussian distribution with mean $\mu$ and co-variance $\Sigma$.
For a function $\sigma(.)$, we use $\dot{\sigma}(.)$ to represent its derivative.
We use $\textbf{1}_{n\times n}$ for the $n\times n$  matrix of ones, $I_n$ for identity matrix of size $n\times n$, $\expectation{}{.}$ for expectation, $\norm{.}_F$ denotes  Frobenius norm, and $[d]=\{1,2,\ldots,d\}$.

\section{NTK Captures the Behaviour of Vanilla GCN}
\label{sec:vanilla_gcn}
We consider the problem of node classification in graphs in a semi-supervised setting,\footnote{More precisely, transductive setting as we assume all features are available during training at the same time. 
} where the labels are observed only for a subset of the nodes.
We start with the formal setup and NTK derivation for the standard (vanilla) GCN proposed in \citet{kipf2017semi}.

\textbf{Formal Setup.}
Given a graph with $n$ nodes and a set of node features $\{x_i\}_{i=1}^n \subset \sR^f$, we may assume without loss of generality that the set of observed labels $\{y_i\}_{i=1}^m$ correspond to first $m$ nodes. We consider a binary classification problem in this paper to simplify the NTK derivation, that is  $y_i \in \{\pm 1\}$, but this could be extended to multi-class problems. The goal is to correctly predict the $n-m$ unknown labels $\{y_i\}_{i=m+1}^n$. 
We represent the observed labels of $m$ nodes as $Y \in \{\pm 1\}^{m \times 1}$, and the node features as $X\in\sR^{n\times f}$ with the assumption that entire $X$ is available during training.  
We define $S$ to be the graph diffusion operator. The analysis holds for any diffusion $S$, but for simulations, we consider the symmetric degree normalized diffusion $S:=(D+I_n)^{-\frac{1}{2}}(A+I_n)(D+I_n)^{-\frac{1}{2}}$ where $A$ is the adjacency matrix and $D$ is the degree matrix.
We define the GCN of depth $d$ as,
\begin{align}
    F_W(X,S) := \sigmoid \lp {\sqrt{\dfrac{c_\sigma}{h_d}} S \ldots \sigma \lp \sqrt{\dfrac{c_\sigma}{h_1}} S \sigma \lp {S X W_1} \rp W_2 \rp \ldots W_{d+1}} \rp \label{eq:gcn}
\end{align}
where $W:=\{W_i \in \sR^{h_{i-1} \times h_i}\}_{i=1}^{d+1}$ is the set of learnable weight matrices with $h_0=f$ and $h_{d+1}=1$, and $\sigmoid : \sR \rightarrow \lp -1,+1 \rp$ is re-scaled sigmoid since we consider binary node classification with labels in $\{\pm 1\}$, $h_i$ is the size of layer $i \in [d]$ and $\sigma: \sR \rightarrow \sR$ is the point-wise activation function.
We initialise all the weights to be i.i.d $\mathcal{N}(0,1)$ and optimise it using stochastic gradient descent.
We study the limiting behavior of this network with respect to the width, that is, $h_1,\ldots, h_{d} \rightarrow \infty$.

\begin{remark}[$c_\sigma$]
\label{rem:csig_gcn}
While this setup is similar to \citet{kipf2017semi}, it is important to note that we additionally consider the normalisation $\sqrt{{c_\sigma} / {h_i}}$ for layer $i$ to ensure that the input norm is approximately preserved. 
Here, $c_\sigma$ is a scaling factor to normalize the input in the initialization phase and $\textstyle c_\sigma = \Big( \expectation{u \sim \mathcal{N}(0,1)}{\lp \sigma(u) \rp^2}  \Big)^{-1}$ from \citet{du2019gradient}. 
We discuss the role of this normalisation in Section \ref{sec:norm_role}.
\end{remark}
\subsection{NTK for Vanilla GCN}
We derive the NTK for vanilla GCN by first rewriting $F_W(X,S)$  as defined in \eqref{eq:gcn} using the following recursive definitions: 
\begin{align}
     &g_1 :=SX, \qquad g_i := \sqrt{\dfrac{c_\sigma}{h_{i-1}}} S \sigma(f_{i-1}) ~\forall i \in \{2,\ldots,d+1\},  \qquad f_i := g_i W_i ~\forall i \in [d+1]\nonumber \\
    \text{Output:}& \quad F_W(X,S) := \sigmoid(f_{d+1}), \quad \text{where } \sigmoid(x) := \dfrac{2}{1+ \exp(-x)} - 1  \label{eq:def_gcn}
\end{align}
Using the definitions in \eqref{eq:def_gcn}, the gradient with respect to $W_i$ of node $u$ can be written as 
\begin{align}
\lp \dfrac{\partial F_W(X,S)}{\partial W_i} \rp_u := g_i^T\lp b_i \rp_u\quad\text{with} \qquad \nonumber \\
(b_{i})_u = \begin{cases}
        (\dot{\sigmoid}(f_{d+1}))_u & \text{if $i=d+1$} \\
        \sqrt{\dfrac{c_\sigma}{h_i}} (S)_u^T(b_{d+1})_u W_{d+1}^T  \odot (\dot{\sigma}(f_i))_{u.}& \text{if $i=d$} \\
        \sqrt{\dfrac{c_\sigma}{h_i}} S^T(b_{i+1})_u W_{i+1}^T  \odot (\dot{\sigma}(f_i))_{u.} & \text{if $i<d$}
    \end{cases} \label{eq:def_gcn_grad}
\end{align}
and $b_{d+1} := \dot{\sigmoid}(f_{d+1})$.
We derive the NTK, as defined in \eqref{eq:ntk},  using the recursive definition of $F_W(X,S)$ in \eqref{eq:def_gcn} and its derivative in \eqref{eq:def_gcn_grad}. 
The following theorem defines the NTK between every pair of nodes, and the $n \times n$ NTK matrix can be computed at once, as shown below (proof in appendix). 

\begin{theorem}[NTK for Vanilla GCN]\label{th: NTK for GCN}
For the vanilla GCN defined in \eqref{eq:gcn},
the NTK $\Theta$ is given by 
\begin{align}
     \Theta = \sum_{k=1}^{d+1} \strut\smash{\underbrace{\Big( S \Big( \ldots S \Big( S}_{d+1-k \text{ terms}}} \lp \Sigma_k \odot \dot{E}_k \rp S^T \odot  \dot{E}_{k+1} \Big) S^T \odot \ldots \odot \dot{E}_{d}\Big) S^T \Big) \odot \dot{E}_{d+1} .
    \label{eq:dwi}
\end{align}
Here $\Sigma_i\in\sR^{n \times n}$ is the co-variance between nodes of the layer $f_i$, and is given by
$\Sigma_1 := SXX^TS^T$, $\Sigma_i := S E_{i-1} S^T$ with $E_i:= c_\sigma \mathbb{E}_{f \sim \mathcal{N}(0, \Sigma_{i})}\big[{\sigma(f) \sigma(f)^T}\big]$ and $\dot{E}_i:= c_\sigma \mathbb{E}_{f \sim \mathcal{N}(0, \Sigma_{i})}\big[{\dot\sigma(f) \dot\sigma(f)^T}\big] \, \forall i \in [1,d]$, and $\dot{E}_{d+1} := \expectation{f\sim \mathcal{N}(0,\Sigma_{d+1})}{ \dot{\sigmoid} \lp f \rp \dot{\sigmoid} \lp f \rp^T }$.

Each entry of the expected matrix in \eqref{eq:dwi} can be approximately computed as follows. For $\Delta\in \sR^{2\times2}$,
\begin{align}
    \expectation{(p,q) \sim  \mathcal{N}(0, \Delta)}{\dot{\sigmoid} \lp p \rp \dot{\sigmoid} \lp q \rp}
    &= \dfrac{1}{4} - \dfrac{\Delta_{00} + \Delta_{11}}{16} + \dfrac{\Delta_{00} \Delta_{11} + 2 \Delta_{01}^2}{64} + \dfrac{\Delta_{00}^2 + \Delta_{11}^2}{32} + \dfrac{\epsilon^3}{16} \nonumber
\end{align}
for $\vert \epsilon \vert \leq \max{ \{\Delta_{00},\Delta_{11} \} }$.

\end{theorem}
\textbf{Inference using NTK.} 
The NTK matrix $\Theta\in\sR^{n\times n}$ defines the pairwise kernel among all labeled and unlabeled nodes, where each entry $\Theta_{pq}$ represents the kernel between nodes (or features) $x_p$ and $x_q$.
For inference, consider the sub-matrix $\Theta_l \in \sR^{m\times m}$  that consists of the kernel computed between all pairs of labeled nodes, and $\Theta_u\in\sR^{(n-m)\times m}$ that consists of the kernel computed between all pairs of unlabeled and labeled nodes.
In the case of squared loss minimisation by stochastic gradient descent with infinitesimally small learning rate $\eta \rightarrow 0$, the training dynamics resemble kernel regression \citep{arora2019exact}.
Hence, the labels for unlabeled nodes $Y_u$ can be inferred as 
\begin{align}
    Y_u = \Theta_u \Theta_l^{-1} Y ~\in\sR^{n-m} \label{eq:ntk_inference}
\end{align}
which, when thresholded entry-wise at $0$, yields the class prediction for unlabeled nodes.

The NTK derived in \eqref{eq:dwi} holds for vanilla GCN with arbitrary activation function in \eqref{eq:gcn}.
Since the focus of this work is explaining the empirical performance trends of GCNs, we focus on specific activation functions that fix the network architecture allowing the NTK to be evaluated exactly.
We first consider a linear activation, that results in the SGC network \citep{wu2019simplifying}, and derive the NTK as follows. 
\begin{corollary}[Linear GCN]\label{cor:linear_gcn}
Consider $\sigma(x) :=x$ in $F_W(X,S)$, then $E_i = c_\sigma \Sigma_i$ and $\dot{E}_i=c_\sigma {\bf{1}}_{n\times n}$ in Theorem~\ref{th: NTK for GCN}, resulting in the following NTK 
\begin{align*}
 \Theta = c_\sigma^{d} S^{d+1}XX^T S^{(d+1)T} \odot \, \expectation{f\sim \mathcal{N}(0,\Sigma_{d+1})}{ \dot{\sigmoid} \lp f \rp \dot{\sigmoid} \lp f \rp^T }.
\end{align*}
where the last expectation is approximated as in Theorem~\ref{th: NTK for GCN}.
The natural choice of normalisation constant $c_\sigma$ is $c_\sigma=1$ based on Remark~\ref{rem:csig_gcn}.
\end{corollary}

Considering a non-linear network with ReLU activation, the NTK can be computed as shown below. 

\begin{corollary}[ReLU GCN]\label{cor:relu_gcn}
Consider $\sigma(x):=\text{ReLU}(x)$ in $F_W(X,S)$. 
The NTK kernel is computed as in \eqref{eq:dwi}, where given $\Sigma_i$ at each layer, one can evaluate the entries of $E_i$ and $\dot{E}_i  \, \forall i \in [1,d]$ using a result from \citet{bietti2019inductive} as 
\begin{align}
    \Big( E_i \Big)_{pq} = \dfrac{c_\sigma}{2} \sqrt{\lp\Sigma_i\rp_{pp} \lp\Sigma_i\rp_{qq}} \, \, \kappa_1 \lp \dfrac{\lp\Sigma_i\rp_{pq}}{\sqrt{\lp\Sigma_i\rp_{pp} \lp\Sigma_i\rp_{qq}}} \rp \text{ and } 
    \lp \dot{E}_i \rp_{pq} = \dfrac{c_\sigma}{2} \kappa_0 \lp \dfrac{\lp\Sigma_i\rp_{pq}}{\sqrt{\lp\Sigma_i\rp_{pp} \lp\Sigma_i\rp_{qq}}} \rp, \label{eq:exp_relu_der}
\end{align}
where $\kappa_0(x) := \dfrac{1}{\pi}\lp \pi - \text{arccos}\lp x \rp \rp $ and $\kappa_1(x) := \dfrac{1}{\pi} \lp x \lp \pi - \text{arccos}\lp x \rp  \rp + \sqrt{1 - x^2} \rp$.
Based on Remark~\ref{rem:csig_gcn}, the natural choice for
normalisation constant $c_\sigma$ is $c_\sigma=2$. 
\end{corollary}


\subsection{Empirical Analysis of Depth}
Many studies have shown that the performance of vanilla GCN drastically drops with depth due to the over smoothing effect of convolutional layers \citep{li2018deeper, kipf2017semi, chenWHDL2020gcnii}. 
To validate it, we empirically study the performances of GCN and its NTK counterpart.
We use Tesla K80 GPU with 12GB memory from Google Colab to obtain all our experimental results.

\textbf{Experimental Setup.} 
We evaluate the performances of linear and ReLU GCN as stated in Corollary~\ref{cor:linear_gcn} and \ref{cor:relu_gcn}, respectively, and their corresponding NTKs for different depths $d=\{1,2,4,8,16,32,64,128\}$.
We fix the size of hidden layers $h_i$ to be the same across all layers to reduce the number of hyper-parameters. 
We consider a range of learning rates $\eta = \{10^{-2},10^{-3},10^{-4}\}$, different size of the hidden layers $h_i = \{16,64,128,256\}$ and report the best performance among the different $\eta$ and size $h_i$ over $10,000$ epochs.
It is important to note that the chosen learning rates are in accordance to the theoretical analysis, that is, $\eta\to0$.
We conduct the experiments with three datasets, namely \emph{Cora} \citep{mccallum2000automating}, \emph{Citeseer} \citep{giles1998citeseer} and \emph{WebKB} \citep{craven1998learning}.
Since the datasets are for multi-class node classification, we combine the classes into two groups to fit our problem in focus -- binary node classification. 
The choice of class grouping is decided by comparing the performances of different groupings and ensuring that the two groups are approximately equal sized.
Appendix B includes detailed discussion on the datasets and grouping of the classes.

\begin{figure}
    \centering
    \includegraphics[width = \linewidth]{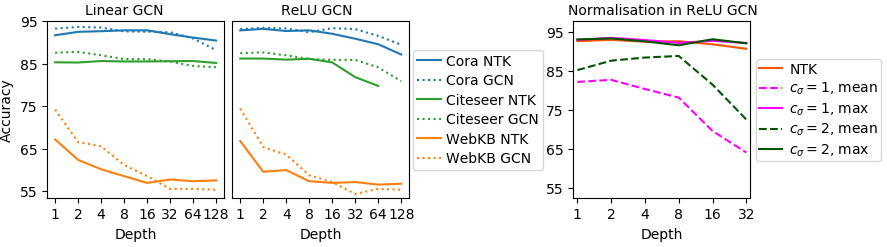}
    \caption{\textbf{(left/middle)} Performance of NTK vs GCN in linear and non-linear architectures. 
    The performance trend of NTK matches the best performance of its corresponding GCN in both the architectures.
    \textbf{(right)} Impact of normalisation in  ReLU GCN evaluated on Cora dataset. The correct choice of normalisation ($c_\sigma=2$ in this case) stabilises the training of GCN even in higher depths, and enables identifying the hyper-parameters in lesser time compared to the unnormalised GCN.
    }
    \label{fig:vanilla_gcn_all}
\end{figure}

\textbf{NTK captures the performance trend of GCN.}
The best performance of GCN decreases with depth in both linear and non-linear architectures, as observed in other papers.
This trend in the best performance is also confirmed in NTK and thus making it a suitable method to analyse finite width GCN, despite the fact that the actual performance of the NTK is usually worse than the corresponding GCN. 
The left plot of Figure~\ref{fig:vanilla_gcn_all}\footnote{NTK for Citeseer faced out-of-memory issue for depth $d=128$ in some cases (can also be seen in Figure~\ref{fig: skip}).}  shows the best performance of both the GCN architectures with its NTK counterpart. 
While there is a drop in the best performance in both the GCNs and the corresponding NTKs, the drop is not as drastic as it has been reported in other papers.
This is due to two factors: first, unlike the previous works that evaluated the performance for a fixed network parameterisation, we allow the size of hidden layers to be chosen as a hyper-parameter.
We found that increasing the network size $h_i$ and/or decreasing the learning rate $\eta$ can reduce the performance drop with depth.
For instance, in \emph{Cora} the best performing network of depth $2$ is achieved with $h_i=16$ and $\eta=10^{-2}$, whereas, $h_i$ has to be increased to $256$ and $\eta$ has to be reduced to $10^{-4}$ for depth $128$ to achieve similar performance.
Second, we identify that the normalisation constant $c_\sigma$ plays a crucial role in stabilising the GCN training.
The right plot of Figure~\ref{fig:vanilla_gcn_all} shows the average and the best performance of unnormalised ($c_\sigma=1$) and correctly normalised ($c_\sigma=2$) ReLU GCN for a fixed parameterisation with $h_i=16$ and trained with learning rate $\eta=\{10^{-2},10^{-3},10^{-4}\}$ over $10,000$ epochs. 
While the average performance of both unnormalised and normalised GCNs shows a drastic drop, correct normalisation enables the network to learn faster and achieve best results. 
Further detailed discussion on the role of normalisation constant $c_\sigma$ is provided in Section~\ref{sec:norm_role}.

\section{NTK - Surrogate for GCN to Analyse Skip Connections}
\label{sec:skip_gcn}
Skip connections \citep{chenWHDL2020gcnii,kipf2017semi} are one way to overcome the performance degradation with depth in GCNs, but little is known about the effectiveness of different forms of available skip connections.
Inspired by the observation of the previous section that the NTK is a hyper-parameter free model that captures the trends of GCNs, we propose NTK as an efficient surrogate for GCN, and we investigate different skip connections for GCN in detail in this section.
We consider two  formulations of skip connections with two variants each that are described in subsequent sections.
To facilitate skip connections, we need to enforce constant layer size, that is, $h_i=h_{i-1}$.
Therefore, we transform the input layer to $H_0$ of size $n \times h$ where $h$ is the hidden layer size. This transformation is necessary as otherwise we would have to assume $h_i = f~\forall i \in[d]$ and $h_i \rightarrow \infty$ would not be possible. 
For this work, we do not consider this transformation as a learnable parameter in the network. As we consider constant layer size, the NTKs are derived considering $h \rightarrow \infty$.
We first define a skip connection related to the one in \citet{kipf2017semi}, where the skip connection is added to the features before convolution (we refer to it as pre-convolution or Skip-PC).

\begin{definition}[Skip-PC]
\label{def:pc}
In a Skip-PC (pre-convolution) network, the transformed input $H_0$ is added to the hidden layers before applying the diffusion, leading to the changes in the recursive definition of \eqref{eq:def_gcn} with $g_1:= \sqrt{\frac{c_\sigma}{h}} \, S \sigma_s(H_0)$ and
\begin{align}
    g_i := \sqrt{\dfrac{c_\sigma}{h}}  S \lp \sigma \lp f_{i-1} \rp + \sigma_s \lp H_0 \rp \rp ~\forall i \in \{2,\ldots,d+1\},  \quad f_i := g_i W_i ~\forall i \in [d+1]
    \label{eq:skip1}
\end{align}
where $\sigma_s(.)$ can be linear or ReLU accounting for two different skip connections. 
\end{definition}

We refer to the network with linear $\sigma_s(.)$ and ReLU $\sigma_s(.)$ as Linear Skip-PC and ReLU Skip-PC, respectively.
The above definition deviates from \citet{kipf2017semi} in the fact that we skip to the input layer instead of the previous layer.
This particular change helps in evaluating the importance of graph information in a dataset which we discuss in the following section.
We also consider a skip connection similar to the one described in \citet{chenWHDL2020gcnii}.

\begin{definition}[Skip-$\alpha$] 
\label{def:alp}
Given an interpolation coefficient $\alpha\in(0,1)$ and a function $\sigma_s(\cdot)$, a Skip-$\alpha$  network is defined such that the transformed input $H_0$ and the hidden layer are interpolated linearly, which changes the recursive definition in \eqref{eq:def_gcn} as $g_1:= \sqrt{\frac{c_\sigma}{h}} \lp (1-\alpha) S\sigma_s(H_0) + \alpha \sigma_s(H_0) \rp$ and
\begin{align}
    g_i := \sqrt{\dfrac{c_\sigma}{h}} \lp \lp 1- \alpha \rp S \sigma \lp f_{i-1} \rp + \alpha \, \sigma_s \lp H_0 \rp \rp ~\forall i \in \{2,\ldots,d+1\}, \,\,\,\, f_i := g_i W_i ~\forall i \in [d+1]
    \label{eq:skip2}
\end{align}
\end{definition}

Similar to Skip-PC, $\sigma_s(.)$ can be linear or ReLU accounting for two different skip connections.
We refer to the network with linear $\sigma_s(.)$ and ReLU $\sigma_s(.)$ as Linear Skip-$\alpha$ and ReLU Skip-$\alpha$, respectively.
\citet{chenWHDL2020gcnii} recommends the choices for $\alpha$ as $0.1$ or $0.2$. 

\begin{remark}[Change of the normalization factor $c_\sigma$ due to Skip connections]
\label{rem:csig_skip}
Note that the normalisation constant $c_\sigma$ for GCN with skip connections is not the same as defined in Remark~\ref{rem:csig_gcn} of vanilla GCN, since we add the transformed input to the hidden layers.
Intuitively,  $c_\sigma<1$ as the norm of the hidden layers would increase otherwise due to the added term. 
We derived $c_\sigma$ specifically for non-linear GCN with $\sigma(x) := \text{ReLU}(x)$, and it is $\simeq 0.67$.
Refer to Appendix A for the proof.
\end{remark}

\subsection{NTK for GCN with Skip Connections}
We derive NTKs for the skip connections --  Skip-PC and Skip-$\alpha$.
Both the NTKs maintain the form presented in Theorem~\ref{th: NTK for GCN} with the following changes to the co-variance matrices. Let $\Tilde{E}_0 = \expectation{f \sim \mathcal{N}(0, \Sigma_0)}{\sigma_s(f)\sigma_s(f)^T}$.
\begin{corollary}[NTK for Skip-PC] 
\label{cor:pc}
The NTK for an infinitely wide Skip-PC network is as presented in Theorem~\ref{th: NTK for GCN} where $E_i$ is defined as in the theorem, but $\Sigma_i$ is defined as
\begin{align}
    \Sigma_0 := XX^T, \qquad \Sigma_1 := S\Tilde{E}_0S^T \qquad \text{and} \qquad \Sigma_i := S E_{i-1} S^T + \Sigma_1. \label{eq:ntk_skip1}
\end{align}
\end{corollary}

\begin{corollary}[NTK for Skip-$\alpha$] 
\label{cor:alp}
The NTK for an infinitely wide Skip-$\alpha$ network is as presented in Theorem~\ref{th: NTK for GCN} where $E_i$ is defined as in the theorem, but $\Sigma_i$ is defined with $\Sigma_0 := XX^T$,
\begin{align}
     \Sigma_1 &:= \lp 1-\alpha \rp^2 SE_0S^T + \alpha \lp 1-\alpha \rp \lp SE_0 + E_0S^T \rp + \alpha^2 E_0 \nonumber \\
     \Sigma_i &:= (1-\alpha)^2 SE_{i-1}S^T + \alpha^2 \Tilde{E_0}. \label{eq:ntk_skip2}
\end{align}
\end{corollary}
Both Corollary \ref{cor:linear_gcn} and \ref{cor:relu_gcn} for linear and ReLU activations, respectively, hold for the derived NTKs corresponding to Skip-PC and Skip-$\alpha$.

\subsection{Empirical Analysis}
Despite studies \citep{chenWHDL2020gcnii,kipf2017semi} showing that having skip-connections gives a significant performance advantage, there is no clear way to choose one formulation of the skip connection over others.
This practical problem can again be seen in the NTK setting as the derived NTKs have similar structure except the co-variance between the nodes, thus making it difficult to compare analytically.
Therefore, we empirically study the performance of different NTKs in order to determine the preferred formulation, thereby avoiding computational intensive hyper-parameter tuning.
In addition, we show that the NTK corresponding to Skip-$\alpha$ can be used for assessing the relevance of structure and feature information of graph in a dataset.
We study the non-linear ReLU GCN with the discussed skip connections, that is, $\sigma(.):=\text{ReLU}$ in \eqref{eq:gcn} empirically.

\begin{figure}
    \centering 
    \includegraphics[width = \linewidth]{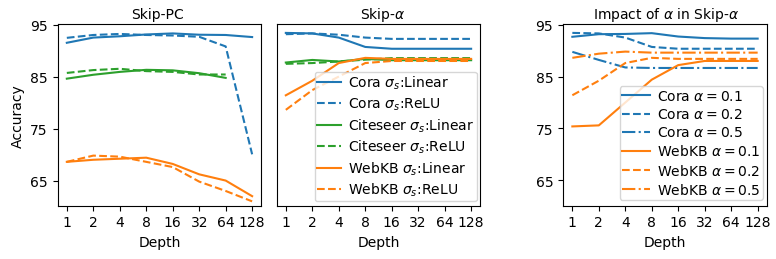}
    \caption{\textbf{(left/middle)} Performance of NTKs corresponding to the different skip connections where
     Skip-$\alpha$ is plotted for $\alpha=0.2$. \textbf{(right)} Impact of $\alpha$ in Skip-$\alpha$ evaluated on Cora and WebKB datasets.}
    \label{fig: skip}
\end{figure}

\textbf{Experimental setup.}
We evaluate the performance of NTKs corresponding to GCNs with skip connections for different depths $d=\{1,2,4,8,16,32,64,128\}$ using non-linear activation $\sigma(x):=\text{ReLU}(x)$ for the GCNs.
The linear transformation of the input $X$ is done by $H_0 = XT$ where $T$ is a $f\times h$ matrix and each entry is sampled from $\mathcal{N}(0,1)$.
The interpolation coefficient $\alpha$ in Skip-$\alpha$ is chosen to be $\{0.1,0.2,0.5\}$.
NTKs for all the formulations of skip connections discussed in the previous section are evaluated on different datasets, namely \emph{Cora}, \emph{Citeseer} and \emph{WebKB}.
Figure~\ref{fig: skip} shows the empirical observations. Refer to Appendix B for more details.

We validate the expected performance advantage of GCN with skip connections over vanilla GCN, more precisely their NTK counterparts,
and observe the following main findings.

\textbf{Non-linear $\sigma_s$ and shallow net for GCNs with skip connection.}
Empirical analysis reveals a distinct behavior of skip connections with $\sigma_s(.)$ being linear and ReLU, which is illustrated in the left plot of Figure~\ref{fig: skip}.
We observe that the performance of both Skip-PC and Skip-$\alpha$ is not optimal at deeper depths, and hence we restrict our focus to shallow depths.
In the case of shallow depths, we find that using non-linear ReLU $\sigma_s(.)$ in both Skip-PC and Skip-$\alpha$ produces the best performance.
Although ReLU Skip-$\alpha$ initially falls short of its counterpart Linear Skip-$\alpha$, it eventually outperforms or performs as good as Linear Skip-$\alpha$, thus favoring ReLU $\sigma_s(.)$.
In addition, this experiment also validates the general practice of using shallow nets for GCNs.
Consequently, we propose skip connections with ReLU $\sigma_s(.)$ and using shallow nets to achieve the best performance in practice.

\textbf{NTK as a model to assess relevance of structure and feature information of graphs.}
In the left plot of Figure~\ref{fig: skip}, we notice that the performance of Skip-$\alpha$ on WebKB improved significantly as compared to Skip-PC and moreover, its performance continued to improve with depth, which is in contrast to other datasets.
We further investigate this by analysing the interpolation coefficient $\alpha$, and the corresponding results on Cora and WebKB datasets are shown in the right plot of Figure~\ref{fig: skip}.
Large value of $\alpha$ in Skip-$\alpha$ implies that more importance is given to feature information than the structural information of the graph. 
Therefore, from the figure, we infer that the structural information is not as important as the feature information for WebKB which is in contrast to Cora.
Besides, NTK is a ready-to-use model without the need for hyper-parameter tuning.
As a result, we propose NTK corresponding to Skip-$\alpha$ as a stand-alone model to determine the relative importance of structure and feature information in tasks where GCNs are employed.


\section{Role of Normalisation in GCN}
\label{sec:norm_role}
\begin{figure}
    \centering
    \includegraphics[width = \linewidth]{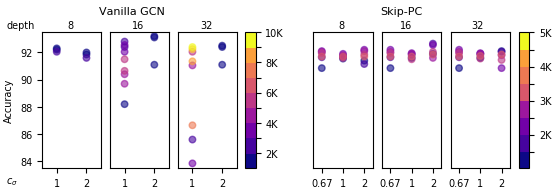}
    \caption{Role of normalisation $c_\sigma$ in vanilla GCN and Skip-PC as defined in \ref{eq:skip1}. The colorbar represents the number of epochs. The correct choice of $c_\sigma$ stabilises the training of GCN even in higher depths in vanilla GCN.}
    \label{fig:normalisation_effect}
\end{figure}

In Section~\ref{sec:vanilla_gcn}, we discussed that the performance drop with depth in vanilla GCN can be reduced by varying the size of the hidden layers $h_i$ rather than fixing the network parameterisation as done in other works.
The main difference between our theoretical setup for infinite width GCN and the practical finite width GCN is the normalisation $\sqrt{{c_\sigma} / {h_i}}$.
Practical networks generally ignore the normalisation factor and rely on weight initialisation and optimisation algorithm to stabilise the training.
Intrigued by this, we investigate the role of normalisation applied to each layer by fixing the network parameterisation in vanilla GCN and Skip-PC empirically.
Figure~\ref{fig:normalisation_effect} illustrates this for different $c_\sigma=\{0.67,1,2\}$ and depths $d=\{8,16,32\}$ on \emph{Cora} dataset.
The considered architectures have non-linear activation, that is, $\sigma(x):=\text{ReLU}(x)$ and we fix the network parameterisation in both the cases.
Different colors in the plot represent the epoch at which the performance is achieved.
The correct choice of $c_\sigma$ is $2$ for ReLU in vanilla GCN (Corollary~\ref{cor:relu_gcn}) and $0.67$ for Skip-PC (Remark~\ref{rem:csig_skip}).

We make the following observation. 
In the case of vanilla GCN, it is clear that the best performance is achieved in almost the same number of epochs across all the depths for the correct choice of $c_\sigma=2$. Moreover, the decrease in the performance for deeper networks is not significant.
Also we need to train the network longer for $c_\sigma=1$ to achieve similar performance of the network with correct normalisation ($c_\sigma=2$) as we increase the depth.
Thus, normalisation plays a crucial role in stabilising the training of vanilla GCN especially in higher depths.
In Skip-PC, the performance of the network is not significantly affected by $c_\sigma$.
This is because the residual connection ensures that the hidden layer norm is approximately equal to the input norm, and thus $c_\sigma$ is not as relevant as it is in vanilla GCN case.
Therefore, in practice, the absence of this normalisation in vanilla GCN explains the reported drastic degradation in performance with depth in the existing literature.

\section{Convergence of NTK with depth}
\label{sec:convergence}

\begin{figure}
    \centering
    \includegraphics[width = \linewidth]{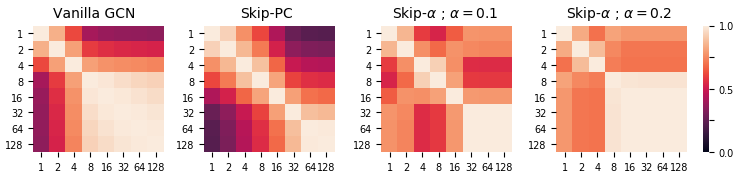}
    \caption{Convergence of NTK with depth for all the discussed ReLU architectures, evaluated on Cora dataset. The plots show perfect alignment of NTKs for higher depths in GCNs with skip connections.}
    \label{fig:ntk_heatmap}
\end{figure}

In Figure~\ref{fig: skip}, we observe that the performance of NTKs corresponding to GCNs with skip connections does not change significantly beyond a certain depth.
We investigate this behaviour of the NTK further by measuring the amount of change between NTKs of different depths. To this end, we consider
the alignment between the NTKs in the eigenspace following \citet[Section 4.2]{fowlkes2004spectral}.
Formally, let $\Theta_i$ and $\Theta_j$ be the NTK of depth $i$ and $j$, respectively, and $U_i^{(k)}$ and $U_j^{(k)}$ be the matrix of $k$ leading eigenvectors of $\Theta_i$ and $\Theta_j$, respectively, then the alignment between $\Theta_i$ and $\Theta_j$ is computed by $a = \frac{1}{k}\norm{U_i^{(k)^T}U_j^{(k)}}_F^2$,
where $a\in[0,1]$ with $a=1$ indicating perfect alignment.
Figure~\ref{fig:ntk_heatmap} shows the alignment of the NTKs for the discussed non-linear ReLU architectures ($\sigma(.):=\text{ReLU}$ in \eqref{eq:gcn}), evaluated on \emph{Cora} dataset. 
Similar pattern is observed in other datasets as well (Appendix B).

\textbf{The learning happens in shallow depth.}
The different alignment plots illustrate the general influence of depth in GCN. 
We observe significant changes in the alignment between NTKs of shallow depths indicating that this is the important part where learning happens.
Since the NTKs for both vanilla GCN and GCN with skip connections converge with depth, it is clear that deep GCNs have no advantage or in other words, no new information is learned at deeper depths.

\textbf{Influence of Skip connection.}
In addition, we observe that the NTKs reach almost perfect alignment with depth for GCNs with skip connection, suggesting that the networks reached saturation in learning as well. We can further distinguish the presented skip connections: overall Skip-PC has slow convergence most likely because the skip connection facilitates learning; 
Skip-$\alpha$ converges fast and as discussed in Section~\ref{sec:skip_gcn}, we observe the influence of $\alpha$ in the learning depending on the dataset. 

\section{Conclusion}
\label{sec:conclusion}
In this work, we derive NTKs for semi-supervised GCNs, including different formulations of skip connections. 
The deterministic hyper-parameter free nature of NTK makes it preferable over its neural network counterpart since it captures the behaviour of the networks very well, as demonstrated in our experiments.
With the support of our empirical results and the findings from \citet{du2019graph} that the NTK for supervised GCN outperforms the neural network, we expect the NTKs for semi-supervised models to perform competitively against the respective neural networks.
Nonetheless, the primary goal of our work is to use NTK to advance our understanding of GCN, particularly on the impact of depth.
%
In addition, we suggest NTK as a surrogate to study variants of GCNs.
From our surrogate analysis, we propose the NTK corresponding to the skip connection Skip-$\alpha$ as an efficient ready-to-use off-the-shelf model to determine the relative importance of structure and feature information in graphs, which we believe to be of great practical value.
There is a possibility of expanding the usage of NTK surrogates to analyse robustness or explainability of GCNs, or other contexts that involve repeated training of networks.
Another direction of research is to incorporate practical considerations of network architecture in the NTK derivation. 
The present paper allows sigmoid functions in the output layer, which is included through a Taylor expansion. It would be also interesting to derive NTKs considering approximations for softmax, max-pooling, dropout or batch normalisation, and use the NTKs to analyse the impact of these techniques on network performance.



\nocite{domingos2020every,lee2018deep,zhou2020graph}

\clearpage
\bibliographystyle{plainnat}
\bibliography{ref.bib}

\clearpage
\section{Proofs of NTKs for GCN and GCN with Skip Connections}
\label{app:proofs}
We provide proofs of Theorem~\ref{th: NTK for GCN} and all corollaries with additional empirical results in this section.

\subsection{Proof of NTK for Vanilla GCN (Theorem~\ref{th: NTK for GCN})}
\textbf{Co-variance between Nodes.} We will first derive the co-variance matrix of size $n \times n$ for each layer comprising of co-variance between any two nodes $p$ and $q$. 
The co-variance between $p$ and $q$ in $f_1$ and $f_i$ are derived below.
We denote $p$-th row of matrix $M$ as $M_{p.}$ throughout our proofs.

\begin{align}
    \expectation{}{ \lp f_1 \rp_{pk} \lp f_1 \rp_{q k^\prime}} &= \expectation{}{ \lp g_1 W_1 \rp_{pk} \lp g_1 W_1 \rp_{q k^\prime}} \nonumber \\
    &= \expectation{}{ \sum_{r=1}^{h_0}  \lp g_1 \rp_{pr} \lp W_1 \rp_{rk} \sum_{s=1}^{h_0} \lp g_1 \rp_{qs} \lp W_1 \rp_{sk^\prime}} \stackrel{\lp W_1 \rp_{xy} \sim \mathcal{N}(0,1)}{=} 0 \quad \text{; if $r \ne s$ or $k \ne k^\prime$} \nonumber \\ 
    \expectation{}{ \lp f_1 \rp_{pk} \lp f_1 \rp_{q k}} &\stackrel[k=k^\prime]{r=s}{=} \expectation{}{ \sum_{r=1}^{h_0}  \lp g_1 \rp_{pr} \lp g_1 \rp_{qr} \lp W_1 \rp_{r k}^2} \nonumber \\
    &\stackrel{\lp W_1 \rp_{xy} \sim \mathcal{N}(0,1)}{=} \sum_{r=1}^{h_0}  \lp g_1 \rp_{pr} \lp g_1 \rp_{qr} = \linnerprod \lp g_1 \rp_{p.} , \lp g_1 \rp_{q.} \rinnerprod \label{eq:Ef1} \\ 
    \expectation{}{ \lp f_i \rp_{pk} \lp f_i \rp_{q k}} &\stackrel[k = k^\prime]{r=s}{=} \expectation{}{ \sum_{r=1}^{h_{i-1}}  \lp g_i \rp_{pr} \lp g_i \rp_{qr} \lp W_i \rp_{r k}^2} \nonumber \\
    &\stackrel{\lp W_i \rp_{xy} \sim \mathcal{N}(0,1)}{=} \sum_{r=1}^{h_{i-1}}  \lp g_i \rp_{pr} \lp g_i \rp_{qr} = \linnerprod \lp g_i \rp_{p.} , \lp g_i \rp_{q.} \rinnerprod \label{eq:Efi}
\end{align}

\begin{align}
    \eqref{eq:Ef1} : \quad \linnerprod \lp g_1 \rp_{p.} , \lp g_1 \rp_{q.} \rinnerprod &= \linnerprod \lp S X \rp_{p.} , \lp S X \rp_{q.} \rinnerprod = S_{p.} X X^T S_{.q}^T = \lp \Sigma_{1} \rp_{pq}  \label{eq:sig1} \\
    \eqref{eq:Efi} : \quad  \linnerprod \lp g_i \rp_{p.} , \lp g_i \rp_{q.} \rinnerprod &= \dfrac{c_\sigma}{h_{i-1}} \linnerprod \lp S \sigma(f_{i-1}) \rp_{p.} , \lp S \sigma(f_{i-1}) \rp_{q.} \rinnerprod \nonumber \\
    &= \dfrac{c_\sigma}{h_{i-1}} \sum_{k=1}^{h_{i-1}} \lp S \sigma(f_{i-1}) \rp_{pk} \lp S \sigma(f_{i-1}) \rp_{qk} \nonumber \\
    &\stackrel{h_{i-1} \rightarrow \infty}{=} c_\sigma \expectation{}{ \lp S \sigma(f_{i-1}) \rp_{pk} \lp S \sigma(f_{i-1}) \rp_{qk}} \qquad \text{; law of large numbers} \nonumber \\
    &= c_\sigma \expectation{}{ \lp \sum_{r=1}^n S_{pr} \sigma \lp f_{i-1} \rp_{rk} \rp \lp \sum_{s =1}^n S_{qs} \sigma \lp f_{i-1} \rp_{s k} \rp} \nonumber \\
    &= c_\sigma \expectation{}{ \sum_{r=1}^n \sum_{s =1}^n S_{pr} S_{qs} \sigma \lp f_{i-1} \rp_{rk}  \sigma \lp f_{i-1} \rp_{s k} }  \nonumber \\
    &\stackrel{(a)}{=} \sum_{r=1}^n \sum_{s =1}^n S_{pr} \lp E_{i-1} \rp_{rs} S_{s q}^T = S_{p.} E_{i-1} S_{.q}^T =  \lp \Sigma_{i} \rp_{pq} \label{eq:Sigma_i}
\end{align}
$(a)$: using $\expectation{}{ \lp f_{i-1} \rp_{rk} \lp f_{i-1} \rp_{sk}} = \lp \Sigma_{i-1}\rp_{rs}$ and the definition of $E_{i-1}$ in Theorem~\ref{th: NTK for GCN}. 

\textbf{NTK for Vanilla GCN.}
Let us first evaluate the tangent kernel component from $W_i$ respective to nodes $p$ and $q$.
We first evaluate  $\linnerprod \lp \der{F}{W_i} \rp_p , \lp \der{F}{W_i} \rp_q \rinnerprod$ in the following.
\begin{align}
    \linnerprod \lp \der{F}{W_i} \rp_p , \lp \der{F}{W_i} \rp_q \rinnerprod &= \sum_{r=1,s=1}^{h_{k-1}, h_k} \lp\lp \der{F}{W_i} \rp_p\rp_{rs} \lp \lp \der{F}{W_i} \rp_q  \rp_{rs} \nonumber \\
    &= \sum_{r=1, s=1}^{h_{k-1}, h_k} \lp g_i^T \lp b_i\rp_p \rp_{rs} \lp g_i^T \lp b_i\rp_q \rp_{rs} \nonumber \\
    &= \sum_{r=1, s=1}^{h_{k-1}, h_k} \sum_{a=1, b=1}^{n,n} \lp g_i^T \rp_{ra} \lp\lp b_i \rp_p \rp_{as} \lp g_i^T \rp_{rb} \lp\lp b_i\rp_q \rp_{bs} \nonumber \\
    &= \sum_{s=1}^{h_k} \sum_{a=1, b=1}^{n,n} \dfrac{c_\sigma}{h_k} \lp S^T \lp b_{i+1}\rp_p W_{i+1}^T\rp_{as} \lp \dot{\sigma}\lp F_i \rp \rp_{ar} \lp g_i g_i^T \rp_{ab} \nonumber \\
    &\qquad \qquad \qquad \lp S^T \lp b_{i+1}\rp_p W_{i+1}^T\rp_{br} \lp \dot{\sigma}\lp F_i \rp \rp_{br} \label{eq:GtB} \\
    &= \sum_{r=1, l=1, m=1}^{h_k, h_{k+1}, h_{k+1}} \sum_{a=1, b=1}^{n,n} \dfrac{c_\sigma}{h_k} \lp S^T \lp b_{i+1}\rp_p\rp_{al} \lp W_{i+1}^T\rp_{lr} \lp \dot{\sigma}\lp F_i \rp \rp_{ar} \lp g_i g_i^T \rp_{ab} \nonumber \\
    &\qquad \qquad \qquad \qquad \lp S^T \lp b_{i+1}\rp_p \rp_{bm} \lp W_{i+1}^T\rp_{mr} \lp \dot{\sigma}\lp F_i \rp \rp_{br}\nonumber 
\end{align}
\begin{align}
    &\stackrel[h_{k+1} \to \infty]{h_k \to \infty}{=} c_\sigma \sum_{r=1, l=1}^{h_k, h_{k+1}} \sum_{a=1, b=1}^{n,n} \lp S^T \lp b_{i+1}\rp_p\rp_{al} \lp \dot{\sigma}\lp F_i \rp \rp_{ar} \lp g_i g_i^T \rp_{ab} \lp S^T \lp b_{i+1}\rp_p \rp_{bl} \lp \dot{\sigma}\lp F_i \rp \rp_{br}\nonumber \\
    &= c_\sigma \sum_{ l=1}^{h_{k+1}} \sum_{a=1, b=1}^{n,n} \lp S^T \lp b_{i+1}\rp_p\rp_{al} \lp S^T \lp b_{i+1}\rp_q \rp_{bl} \lp g_i g_i^T \rp_{ab}  \expectation{}{\lp \dot{\sigma}\lp F_i \rp \dot{\sigma}\lp F_i \rp^T\rp_{ab}} \nonumber \\
    &\stackrel{(b)}{=} \sum_{ l=1}^{h_{k+1}}  \lp\lp S^T \lp b_{k=i+1}\rp_p\rp^T  \lp g_i g_i^T \odot \dot{E}_i\rp \lp S^T \lp b_{i+1}\rp_q \rp\rp_{ll} \nonumber \\
    &= \trace(\lp b_{i+1}\rp_p^T S \lp {\Sigma}_i \odot \dot{E}_i \rp S^T \lp b_{i+1}\rp_q) \nonumber \\
    &\stackrel{(c)}{=}\trace(\lp b_{d+1}\rp_p^T S_{u} \lp \ldots S \lp S \lp {\Sigma}_i \odot \dot{E}_i \rp S^T \odot  \dot{E}_{i+1} \rp S^T \odot \ldots \odot \dot{E}_{d}\rp S^T_{v} \lp b_{d+1}\rp_q) \nonumber \\
    &= S_{u.} \lp \ldots S \lp S \lp {\Sigma}_i \odot \dot{E}_i \rp S^T \odot  \dot{E}_{i+1} \rp S^T \odot \ldots \odot \dot{E}_{d}\rp S^T_{q.} \odot \dot{\sigmoid} \lp f_{d+1} \rp \dot{\sigmoid} \lp f_{d+1} \rp^T_{pq} \label{eq:ntk_Wi_pq}
\end{align}
(b): $c_\sigma \expectation{}{\lp \dot{\sigma}\lp F_i \rp \dot{\sigma}\lp F_i \rp ^T\rp_{ab}} = \lp \dot{E}_k \rp_{ab}$. \\
(c): Expanding $b_{i+1}$ will result in the expression similar to \eqref{eq:GtB}, and repeated expansion until $b_{d+1}$. 
The final equation is obtained by substituting $ b_{d+1} = \dot{\sigmoid} \lp f_{d+1} \rp$ from its definition in \eqref{eq:dwi}.

Extending \eqref{eq:ntk_Wi_pq} to all $n$ nodes which will result in $n \times n$ matrix,

\begin{align}
    \linnerprod \der{F}{W_i}, \der{F}{W_i} \rinnerprod &= S \lp \ldots S \lp S \lp {\Sigma}_i \odot \dot{E}_i \rp S^T \odot  \dot{E}_{i+1} \rp S^T \odot \ldots \odot \dot{E}_{d}\rp S^T \nonumber \\
    &\qquad  \odot \dot{\sigmoid} \lp f_{d+1} \rp \dot{\sigmoid} \lp f_{d+1} \rp^T \nonumber \\
    \expectation{W_i}{\linnerprod \der{F}{W_i}, \der{F}{W_i} \rinnerprod} &= S \lp \ldots S \lp S \lp {\Sigma}_i \odot \dot{E}_i \rp S^T \odot  \dot{E}_{i+1} \rp S^T \odot \ldots \odot \dot{E}_{d}\rp S^T \nonumber \\
    &\qquad \odot \expectation{f \sim \mathcal{N} \lp 0, \Sigma_{d+1} \rp}{\dot{\sigmoid} \lp f \rp \dot{\sigmoid} \lp f \rp^T } \label{eq:ntk_Wi}
\end{align}

Finally, NTK $\Theta$ is,
\begin{align}
    \Theta &= \sum_{i=1}^{d+1} \expectation{W_i}{\linnerprod \der{F}{W_i}, \der{F}{W_i} \rinnerprod} \nonumber \\
    &= \lb \sum_{i=1}^{d+1} S \lp \ldots S \lp S \lp {\Sigma}_i \odot \dot{E}_i \rp S^T \odot  \dot{E}_{i+1} \rp S^T \odot \ldots \odot \dot{E}_{d}\rp S^T \rb \nonumber \\
    &\qquad \odot \, \expectation{f\sim \mathcal{N}(0,\Sigma_{d+1})}{ \dot{\sigmoid} \lp f \rp \dot{\sigmoid} \lp f \rp^T }  \label{eq:ntk_final}
\end{align}

We will now compute $\expectation{f\sim \mathcal{N}(0,\Sigma_d)}{ \dot{\sigmoid} \lp f \rp \dot{\sigmoid} \lp f \rp^T }$.
We use Lagrange form of the remainder to approximate the Taylor's expansion for the re-scaled sigmoid function $\sigmoid(.)$ which gives better bound. 
\looseness=-1
\begin{align}
    \sigmoid(x) &= \dfrac{2}{1+ \exp^{-x}} -1 = \dfrac{x}{2} - \dfrac{x^3}{24} + \dfrac{x^5}{240} + \cdots \nonumber \\
    \dot{\sigmoid}(x) &= \dfrac{1}{2} - \dfrac{x^2}{8} + \dfrac{x^4}{48} + \dfrac{x^6 \dot{\sigmoid}^6(\xi) }{6!} \qquad \text{; last term is the Lagrange form of the remainder.} \label{eq:rescaled_sig}
\end{align}
\looseness=-1
To evaluate the expectation of an entry $i,j$ in the matrix $\dot{\sigmoid} \lp f \rp \dot{\sigmoid} \lp f \rp^T$, let us define $\Delta$ as a $2 \times 2$ co-variance matrix as follows, $
\Delta=
  \begin{bmatrix}
    \lp \Sigma_{d+1} \rp_{ii} & \lp \Sigma_{d+1} \rp_{ij} \\
    \lp \Sigma_{d+1} \rp_{ji} & \lp \Sigma_{d+1} \rp_{jj}
  \end{bmatrix}
$

\begin{align}
    \expectation{(x,y) \sim \Delta}{\dot{\sigmoid} \lp x \rp \dot{\sigmoid} \lp y \rp} &\stackrel{\eqref{eq:rescaled_sig}}{=} \expectation{(x,y) \sim \Delta}{\lp \dfrac{1}{2} - \dfrac{x^2}{8} + \dfrac{x^4}{48} + \dfrac{x^6 \dot{\sigmoid}^6(\xi) }{6!} \rp \lp \dfrac{1}{2} - \dfrac{y^2}{8} + \dfrac{y^4}{48} + \dfrac{y^6 \dot{\sigmoid}^6(\xi) }{6!} \rp} \nonumber \\
    &= \dfrac{1}{4} \underset{(x,y) \sim \Delta}{\mathbb{E}}\Big[ 1 - \dfrac{x^2}{4} - \dfrac{y^2}{4} + \dfrac{x^4}{24} + \dfrac{y^4}{24} + \dfrac{x^2 y^2}{16} - \dfrac{x^4y^2}{96} - \dfrac{x^2 y^4}{96} \nonumber \\
    &\qquad \qquad + \dfrac{x^4y^4}{576} + \dfrac{x^6 \dot{\sigmoid}^6(\xi) }{6!} \lp \dfrac{1}{2} - \dfrac{y^2}{8} + \dfrac{y^4}{48} + \dfrac{y^6 \dot{\sigmoid}^6(\xi) }{6!} \rp \Big] \label{eq:exp_der_sig}
\end{align}

\textbf{Compute $\expectation{x \sim \mathcal{N}(0,\lambda^2)}{x^k}$ and $\expectation{(x,y) \sim \mathcal{N}(0,\Delta)}{x^i y^j}$.}
\begin{align}
    \expectation{x \sim \mathcal{N}(0,\lambda^2)}{x^k} &= \dfrac{2 }{\sqrt{2 \pi} \lambda} \int_0^\infty x^k \exp \lp \dfrac{-x^2}{2 \lambda^2} \rp \mathrm{d}x \nonumber \\ 
    &= \dfrac{2 \lambda^k}{\sqrt{2 \pi}} \int_0^\infty t^k \exp \lp \dfrac{-t^2}{2} \rp \mathrm{d}t \qquad \text{; $x = \lambda t \implies \mathrm{d}x = \lambda \mathrm{d}t$} \nonumber \\
    &= \dfrac{2 \lambda^k}{\sqrt{2\pi}} (k-1) \int_0^\infty t^{k-2} \exp \lp \dfrac{-t^2}{2} \rp \mathrm{d}t \nonumber \\
    \text{Thus,} \expectation{x \sim \mathcal{N}(0,\lambda^2)}{x^k} &= (k-1) \lambda^2 \expectation{x \sim \mathcal{N}(0,\lambda^2)}{x^{k-2}} \label{eq:exp_xn}
\end{align}
\looseness=-1
\begin{align}
    \expectation{(x,y) \sim \mathcal{N}(0,\Delta)}{x^i y^j} &= \expectation{(x,y) \sim \mathcal{N}(0,\Delta)}{x^i \lp y \pm \alpha x \rp^j} \qquad ; \text{$\alpha = \dfrac{\expectation{}{xy}}{\expectation{}{x^2}}$ then $x, y- \alpha x$ are independent} \nonumber \\
    &= \expectation{(x,y) \sim \mathcal{N}(0,\Delta)}{x^i \lp \sum_{k=0}^j \comb{j}{k} \lp y - \alpha x \rp^j \lp \alpha x \rp^k \rp} \nonumber \\
    &= \expectation{(x,y) \sim \mathcal{N}(0,\Delta)}{\sum_{k=0}^j \comb{j}{k} \alpha^k \lp y - \alpha x \rp^j  x^{k+i} } \nonumber \\
    &\stackrel{(e)}{=} \sum_{k=0}^j \comb{j}{k} \alpha^k \expectation{(x,y) \sim \mathcal{N}(0,\Delta)}{x^{k+i}} \expectation{(x,y) \sim \mathcal{N}(0,\Delta)}{\lp y - \alpha x \rp^j} \label{eq:exp_xiyj}
\end{align}
$(e)$: $x,(y-\alpha x)$ are independent then $x^a, (y-\alpha x)^b$ are also independent.

Now, we evalute \eqref{eq:exp_der_sig} using \eqref{eq:exp_xn} and \eqref{eq:exp_xiyj} as follows.
\begin{align}
    \eqref{eq:exp_der_sig} &\stackrel{\eqref{eq:exp_xn},\eqref{eq:exp_xiyj}}{=}\dfrac{1}{4} - \dfrac{1}{16} \lp \Sigma_{2_{ii}} + \Sigma_{2_{jj}} \rp + \dfrac{1}{64} \lp \Sigma_{2_{ii}} \Sigma_{2_{jj}} + 2 \Sigma_{2_{ij}}^2 \rp \nonumber \\
    &\qquad + \dfrac{1}{32} \lp \Sigma_{2_{ii}}^2 + \Sigma_{2_{jj}}^2 \rp - \dfrac{1}{128} \lp \Sigma_{2_{ii}}^2 \Sigma_{2_{jj}} + \Sigma_{2_{ii}} \Sigma_{2_{jj}}^2 + 4 \Sigma_{2_{ij}}^2 \Sigma_{2_{ii}} + 4 \Sigma_{2_{ij}}^2 \Sigma_{2_{jj}} \rp \nonumber \\
    &\qquad + \dfrac{1}{768} \lp 3 \Sigma_{2_{ii}}^2 \Sigma_{2_{jj}}^2 + 8 \Sigma_{2_{ij}}^4 + 24 \Sigma_{2_{ij}}^2 \Sigma_{2_{ii}} \Sigma_{2_{jj}} \rp \nonumber \\
    &\qquad+ \expectation{(x,y) \sim \Delta}{\dfrac{x^6 \dot{\sigmoid}^6(\xi) }{6!} \lp \dfrac{1}{2} - \dfrac{y^2}{8} + \dfrac{y^4}{48} + \dfrac{y^6 \dot{\sigmoid}^6(\xi) }{6!} \rp} \nonumber \\
    &\leq \dfrac{1}{4} - \dfrac{1}{16} \lp \Sigma_{2_{ii}} + \Sigma_{2_{jj}} \rp + \dfrac{1}{64} \lp \Sigma_{2_{ii}} \Sigma_{2_{jj}} + 2 \Sigma_{2_{ij}}^2 \rp + \dfrac{1}{32} \lp \Sigma_{2_{ii}}^2 + \Sigma_{2_{jj}}^2 \rp \nonumber \\
    &\qquad  - \dfrac{10}{128} \epsilon^3  + \dfrac{35}{768} \epsilon^4 + \dfrac{15}{720} \epsilon^3 \qquad \text{; $\vert \epsilon \vert \leq \max{\{\Delta_{00},\Delta_{11}\}}, \expectation{}{x^6}=15\Delta_{00}$} \nonumber \\
    &\leq  \dfrac{1}{4} - \dfrac{1}{16} \lp \Sigma_{2_{ii}} + \Sigma_{2_{jj}} \rp + \dfrac{1}{64} \lp \Sigma_{2_{ii}} \Sigma_{2_{jj}} + 2 \Sigma_{2_{ij}}^2 \rp + \dfrac{1}{32} \lp \Sigma_{2_{ii}}^2 + \Sigma_{2_{jj}}^2 \rp + \dfrac{1}{16}\epsilon^3  \label{eq:der_sig}
\end{align}
where $\vert \epsilon \vert \leq \max{\{\Delta_{00},\Delta_{11}\}}$.

We get the NTK in Theorem~\ref{th: NTK for GCN} by putting together \eqref{eq:der_sig} and \eqref{eq:ntk_final}.

\textbf{Corollary 1 (Linear GCN).}
In this case, $\sigma(x):=x$ and so derivative $\dot{\sigma}(x)=1$. Consequently, one can derive $\dot{E}_i = c_\sigma \textbf{1}_{n \times n}$ from its definition. Therefore, we get NTK for linear GCN in Corollary \ref{cor:linear_gcn} by substituting $\dot{E}_i$ in general NTK equation in \eqref{eq:ntk_final}.

\textbf{Corollary 2 (ReLU GCN).}
NTK for ReLU GCN is derived by substituting \eqref{eq:exp_relu_der} in general NTK equation in \eqref{eq:ntk_final} as discussed in the corollary.

\subsection{ Proof of NTK for GCN with Skip Connections (Corollary~\ref{cor:pc} and \ref{cor:alp})}
We derive the NTKs for GCNs with different skip connections, Skip-PC and Skip-$\alpha$ in this section.
We observe that the definitions of $g_i \, \forall i \in [1,d+1]$ are different for GCN with skip connections from the vanilla GCN. 
Despite the difference, the definition of gradient with respect to $W_i$ in \eqref{eq:def_gcn_grad} does not change as $g_i$ in the gradient accounts for the change and moreover, there is no new learnable parameter since the input transformation $H_0 = XT$ where $T_{ij}$ is sampled from $\mathcal{N}(0,1)$ is not learnable in our setting.
Given the fact that the gradient definition holds for GCN with skip connection, the NTK will retain the form from NTK for vanilla GCN as evident from the above derivation.
The change in $g_i$ will only affect the co-variance between nodes.
Hence, we will derive the co-variance matrix for the discussed skip connections, Skip-PC and Skip-$\alpha$ in the following sections.

\textbf{Skip-PC: Co-variance between nodes.} 
The co-variance between nodes $p$ and $q$ in $f_1$ and $f_i$ are derived below.
\begin{align}
    \expectation{}{ \lp f_1 \rp_{pk} \lp f_1 \rp_{q k}} &= \linnerprod \lp g_1 \rp_{p.} , \lp g_1 \rp_{q.} \rinnerprod \nonumber \\
    &= \dfrac{c_\sigma}{h} \linnerprod \lp S \sigma_s(H_0) \rp_{p.} , \lp S \sigma_s(H_0) \rp_{q.} \rinnerprod \nonumber \\
    &= \dfrac{c_\sigma}{h} \sum_{k=1}^{h} \lp S \sigma_s(H_0) \rp_{pk} \lp S \sigma_s(H_0) \rp_{qk} \nonumber \\
    &\stackrel{h \rightarrow \infty}{=} c_\sigma \expectation{}{ \lp S \sigma_s(H_0) \rp_{pk} \lp S\sigma_s(H_0) \rp_{qk}} \qquad ; \text{law of large numbers} \nonumber \\
    &= S_{p.} \Tilde{E}_0 S_{.q}^T  \qquad ; \Tilde{E}_0 = c_\sigma \expectation{f \sim \mathcal{N}(0,XX^T)}{\sigma_s(f) \sigma_s(f)^T} \nonumber\\
    &= \lp \Sigma_{1} \rp_{pq} \label{eq:sig1_pc} 
\end{align}

\begin{align}
    \expectation{}{ \lp f_i \rp_{pk} \lp f_i \rp_{q k}} &= \linnerprod \lp g_i \rp_{p.} , \lp g_i \rp_{q.} \rinnerprod \nonumber \\
    &= \dfrac{c_\sigma}{h} \linnerprod \lp S \lp \sigma(f_{i-1}) + \sigma_s(H_0) \rp  \rp_{p.} , \lp S \lp \sigma(f_{i-1}) + \sigma_s(H_0) \rp  \rp_{q.} \rinnerprod \nonumber \\
    &= \dfrac{c_\sigma}{h} \sum_{k=1}^{h} \lp S \sigma(f_{i-1}) + S \sigma_s(H_0) \rp_{pk} \lp S \sigma(f_{i-1}) + S\sigma_s(H_0) \rp_{qk} \nonumber \\
    &\stackrel{h \rightarrow \infty}{=} c_\sigma \expectation{}{ \lp S \sigma(f_{i-1}) + S\sigma_s(H_0) \rp_{pk} \lp S \sigma(f_{i-1}) + S\sigma_s(H_0) \rp_{qk}} \quad ; \text{law of large numbers} \nonumber \\
    &= c_\sigma \Big[ \expectation{}{\lp S \sigma(f_{i-1}) \rp_{pk} \lp S \sigma(f_{i-1}) \rp_{qk}} + \expectation{}{\lp S \sigma(f_{i-1}) \rp_{pk} \lp S\sigma_s(H_0) \rp_{qk}} \nonumber \\
    &\qquad \qquad + \expectation{}{\lp S \sigma_s(H_0) \rp_{pk} \lp S \sigma(f_{i-1}) \rp_{qk}} + \expectation{}{\lp S\sigma_s(H_0) \rp_{pk} \lp S\sigma_s(H_0) \rp_{qk} } \Big] \nonumber \\
    &= S_{p.} E_{i-1} S_{.q}^T + c_\sigma \expectation{}{\lp S \sigma(f_{i-1}) \rp_{pk} \lp S\sigma_s(XW_0) \rp_{qk}}  \nonumber \\
    &\qquad \qquad + c_\sigma \expectation{}{\lp S\sigma_s(XW_0) \rp_{pk} \lp S \sigma(f_{i-1}) \rp_{qk}} \nonumber \\
    &\qquad \qquad + c_\sigma \expectation{}{ \sum_{r=1}^n \sum_{s =1}^n S_{pr} S_{qs} \sigma_s \lp XW_0 \rp_{rk}  \sigma_s \lp XW_0 \rp_{s k} } \nonumber \\
    &\stackrel{(f)}{=} S_{p.} E_{i-1} S_{.q}^T + c_\sigma S_{p.} \expectation{}{  \sigma_s \lp XW_0 \rp_{rk}  \sigma_s \lp XW_0 \rp_{s k} } S_{.q}^T \nonumber \\
    &= S_{p.} E_{i-1} S_{.q}^T + S_{p.} \Tilde{E}_0 S_{.q}^T = S_{p.} E_{i-1} S_{.q}^T + \lp \Sigma_{1} \rp_{pq} = \lp \Sigma_{i} \rp_{pq} \label{eq:Sigma_i_pc}
\end{align}
$(f)$: $\expectation{}{\lp S \sigma(f_{i-1}) \rp_{pk} \lp S \sigma_s(XW_0) \rp_{qk}}$ and $\expectation{}{\lp S \sigma_s(XW_0) \rp_{pk} \lp S \sigma(f_{i-1}) \rp_{qk}}$ evaluate to $0$ by conditioning on $W_0$ first and rewriting the expectation based on this conditioning.
The terms within expectation are independent when conditioned on $W_0$, and hence it is
$\expectation{W_0}{\expectation{\Sigma_{i-1} \vert W_0}{\lp S \sigma(f_{i-1}) \rp_{pk} \vert W_0} \expectation{\Sigma_{i-1} \vert W_0}{\lp S \sigma_s(XW_0) \rp_{qk} \vert W_0}}$ by taking $h$ in $W_0$ going to infinity first.
Here, $\expectation{\Sigma_{i-1} \vert W_0}{\lp S \sigma_s(XW_0) \rp_{qk} \vert W_0}=0$.

We get the co-variance matrix for all pairs of nodes $\Sigma_1=S \Tilde{E}_0 S^T$ and $\Sigma_i=S E_{i-1} S^T + \Sigma_1 $ from \eqref{eq:sig1_pc} and \eqref{eq:Sigma_i_pc}.

\textbf{Skip-$\alpha$: Co-variance between nodes.}
Let $p$ and $q$ be two nodes and the co-variance between $p$ and $q$ in $f_1$ and $f_i$ are derived below.

\begin{align}
    \expectation{}{ \lp f_1 \rp_{pk} \lp f_1 \rp_{q k}} &= \linnerprod \lp g_1 \rp_{p.} , \lp g_1 \rp_{q.} \rinnerprod \nonumber \\
    &= \dfrac{c_\sigma}{h} \sum_{k=1}^{h} \lp (1-\alpha)S \sigma_s(H_0) + \alpha \sigma_s(H_0) \rp_{pk} \lp (1-\alpha)S \sigma_s(H_0) + \alpha \sigma_s(H_0) \rp_{qk} \nonumber \\
    &\stackrel{h \rightarrow \infty}{=} c_\sigma \expectation{}{ \lp (1-\alpha)S \sigma_s(H_0) + \alpha \sigma_s(H_0) \rp_{pk} \lp (1-\alpha)S \sigma_s(H_0) + \alpha \sigma_s(H_0) \rp_{qk}} \nonumber \\
    &= c_\sigma \Big[ (1-\alpha)^2 \expectation{}{\lp S \sigma_s(H_0) \rp_{pk} \lp S \sigma_s(H_0) \rp_{qk}} \nonumber \\
    &\quad + (1-\alpha)\alpha \lp \expectation{}{\lp S \sigma_s(H_0) \rp_{pk} \lp \sigma_s(H_0) \rp_{qk}}  + \expectation{}{\lp S \sigma_s(H_0) \rp_{qk} \lp \sigma_s(H_0) \rp_{pk}} \rp \nonumber \\
    &\quad + \alpha^2 \expectation{}{\lp \sigma_s(H_0) \rp_{pk} \lp \sigma_s(H_0) \rp_{qk}} \nonumber \\
    &= (1-\alpha)^2 S_{p.} \Tilde{E}_0 S_{.q}^T + (1-\alpha)\alpha \lp S_{p.} \lp \Tilde{E}_0 \rp_{.q} +  \lp \Tilde{E}_0 \rp_{p.} S_{.q}^T \rp + \alpha^2 \lp \Tilde{E}_0 \rp_{pq} \nonumber \\
    &= \lp \Sigma_{1} \rp_{pq} \label{eq:sig1_alp} 
\end{align}

\begin{align}
    \expectation{}{ \lp f_i \rp_{pk} \lp f_i \rp_{q k}} &= \linnerprod \lp g_i \rp_{p.} , \lp g_i \rp_{q.} \rinnerprod \nonumber \\
    &= \dfrac{c_\sigma}{h} \sum_{k=1}^{h} \lp (1-\alpha) S \sigma(f_{i-1}) + \alpha \sigma_s(H_0)  \rp_{pk} \lp (1-\alpha) S \sigma(f_{i-1}) + \alpha \sigma_s(H_0)  \rp_{qk} \nonumber \\
    &\stackrel{h \rightarrow \infty}{=} c_\sigma \expectation{}{ \lp (1-\alpha) S \sigma(f_{i-1}) + \alpha \sigma_s(H_0)  \rp_{pk} \lp (1-\alpha) S \sigma(f_{i-1}) + \alpha \sigma_s(H_0)  \rp_{qk}} \nonumber \\
    &= c_\sigma \Big[ (1-\alpha)^2 \expectation{}{\lp S \sigma(f_{i-1}) \rp_{pk} \lp S \sigma(f_{i-1}) \rp_{qk}}  + \alpha^2 \expectation{}{\lp \sigma_s(H_0) \rp_{pk} \lp \sigma_s(H_0) \rp_{qk} } \nonumber\\
    &\,\, + (1-\alpha)\alpha \lp \expectation{}{\lp S \sigma(f_{i-1}) \rp_{pk} \lp \sigma_s(H_0) \rp_{qk}} + \expectation{}{\lp \sigma_s(H_0) \rp_{pk} \lp S \sigma(f_{i-1}) \rp_{qk}} \rp \Big] \nonumber \\
    &\stackrel{(g)}{=} (1-\alpha)^2 S_{p.} E_{i-1} S_{.q}^T + \alpha^2 \lp \Tilde{E_0} \rp_{pq} = \lp \Sigma_{i} \rp_{pq} \label{eq:Sigma_i_alp}
\end{align}
$(g)$: same argument as $(f)$ in derivation of $\Sigma_i$ in Skip-PC.

We get the co-variance matrix for all pairs of nodes $\Sigma_1= (1-\alpha)^2 S \Tilde{E}_0 S^T + \alpha(1-\alpha) \lp S \Tilde{E}_0 + \Tilde{E}_0 S^T \rp + \alpha^2 \Tilde{E}_0 $ and $\Sigma_i= (1-\alpha)^2 S E_{i-1} S^T + \alpha^2 \Tilde{E}_0 $ from \eqref{eq:sig1_alp} and \eqref{eq:Sigma_i_alp}.

\subsection{Normalisation constant $c_\sigma$ (Remark~\ref{rem:csig_gcn} and ~\ref{rem:csig_skip}).}
We derive the normalisation constant $c_\sigma$ loosely, as the purpose of $c_\sigma$ is to preserve the input norm approximately.
We focus on general form of a network with skip connection (not GCN in particular), where the output vector of size $h$ from any hidden layer $l$ with weight matrix $W \in \mathbb{R}^{h\times h}$ and transformed input vector $X_0$ of size $h$ can be written as $g_l := \sqrt{\dfrac{c_\sigma}{h}} \lp \sigma(W g_{l-1}) + X_0 \rp \in \mathbb{R}^{h \times 1}$.
The role of the normalisation constant $c_\sigma$ is to maintain $\Vert g_l \Vert_2 \simeq \Vert X_0 \Vert_2 $ and is derived as follows. 

\begin{align}
    \Vert X_0 \Vert_2^2 &= \Vert g_l \Vert_2^2 = \dfrac{c_\sigma}{h} \sum_{k=1}^h \lp \sigma(W g_{l-1}) + X_0 \rp_k^2 \nonumber \\
    \Vert X_0 \Vert_2^2 &= c_\sigma \expectation{}{\lp \sigma(W g_{l-1})_k \rp^2 + \lp X_0\rp_k^2 + 2 \sigma(W g_{l-1})_k \lp X_0\rp_k } \qquad ; \text{$h\rightarrow \infty$} \nonumber \\
    \Vert X_0 \Vert_2^2 &= c_\sigma \expectation{u \sim \mathcal{N}(0, \Vert X_0 \Vert^2)}{\lp \sigma(u) \rp^2} + \Vert X_0 \Vert_2^2 \qquad ; \text{$\expectation{}{\sigma(W g_{l-1})_k \lp X_0\rp_k}=0$} \nonumber \\
    c_\sigma &= \lp \expectation{u \sim \mathcal{N}(0, 1)}{\lp \sigma(u) \rp^2} + 1 \rp^{-1} \qquad ; \text{normalised $X_0$} \label{eq:csig}
\end{align}
We use this $c_\sigma$ for GCN with skip connection in our work and it evaluates to $2/3$ for $\sigma(x):=\text{ReLU}(x)$ in GCN as stated in Remark~\ref{rem:csig_skip}.
The evident change for a network without skip connection is to not add $X_0$ in $g_l := \sqrt{\dfrac{c_\sigma}{h}} \sigma(W g_{l-1})$ and consequently by following the proof, we get $c_\sigma =\lp \expectation{u \sim \mathcal{N}(0, 1)}{\lp \sigma(u) \rp^2} \rp^{-1} $ as mentioned in Remark~\ref{rem:csig_gcn}.

\section{Additional Experimental Results}
\label{app:exp}

\subsection{Datasets for binary node classification}
Since the considered datasets \emph{Cora}, \emph{Citeseer} and \emph{WebKB} are for multi-class node classification, we converted the datasets to have binary class by grouping the classes into two sets.
Table~\ref{tab:dataset_binary} shows the label grouping for each dataset and total number of nodes with the grouped labels respectively.
The classes in all the datasets are well balanced and sensible to learn for binary classification problem which is proved from the performance of a simple graph neural network like linear vanilla GCN. The train-test split for each dataset is $708$ and $2000$ nodes for Cora, $312$ and $2000$ for Citeseer, and $377$ and $500$ for WebKB for all the experiments.

\renewcommand{\arraystretch}{1.5}
\begin{table}[h]
    \centering
        \begin{tabular}{ |c|c|c|c|c|c|c|  }
         \hline
         & \multicolumn{2}{|c|}{\textbf{Cora}} & \multicolumn{2}{|c|}{\textbf{Citeseer}} & \multicolumn{2}{|c|}{\textbf{WebKB}} \\
         \hline
         & Class Groups   & \#nodes & Class Groups & \#nodes & Class Groups & \#nodes\\
         \hline
         \textbf{Class 1} & \makecell{\, \\ Neural\_Networks  \\ Theory \\ Probabilistic\_Methods \\ \,}  & 1595 & \makecell{ Agents \\ AI \\ ML }  & 1435 & student  & 415 \\ 
         \hline
         \textbf{Class 2 } & \makecell{\, \\Case\_Based  \\ Rule\_Learning \\ Reinforcement \\ Genetic\_Algorithms \\ \, }  & 1103 & \makecell{ DB \\ IR \\ HCI }  & 1877 & \makecell{ faculty \\ staff \\  course \\  project}  & 462 \\
         \hline
         \textbf{Total} & & 2708 & & 3312 & & 877 \\
         \hline
        \end{tabular}
    \caption{Class grouping in datasets for binary node classification.}
    \label{tab:dataset_binary}
\end{table}

\subsection{Vanilla GCN vs GCN with Skip Connections}

\begin{figure}[b]
    \centering
    \includegraphics{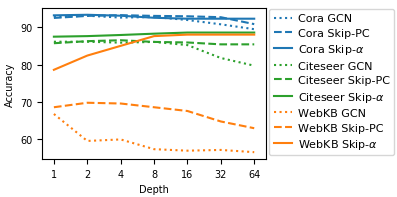}
    \caption{Performance validation of vanilla GCN, Skip-PC and Skip-$\alpha$ with $\sigma(.):=\text{ReLU}$, $\sigma_s(.):=\text{ReLU}$ and $\alpha=0.2$ using the respective NTKs.}
    \label{fig:van_skip}
\end{figure}

We established ReLU GCN is preferred over linear in Section~\ref{sec:vanilla_gcn} and ReLU for the input transformation in Section~\ref{sec:skip_gcn}.
Hence, we focus on $\sigma(.):=\text{ReLU}$ and $\sigma_s(.):=\text{ReLU}$ with $\alpha=0.2$ for Skip-$\alpha$ to validate the performance of vanilla GCN and GCN with skip connections, Skip-PC and Skip-$\alpha$.
We use the respective NTKs to validate the performance.
Figure~\ref{fig:van_skip} shows that GCN with skip connection outperforms vanilla GCN even in deeper depths, and Skip-$\alpha$ gives better performance than Skip-PC with depth. 

\textbf{Note.} In Figure~\ref{fig: skip}, the performance of Skip-PC with ReLU $\sigma_s(.)$ evaluated on Citeseer when depth $=32$ is different from what is plotted due to some numerical precision error.
We evaluated the performance at depth $30,31$ and used it to plot.

\subsection{Convergence of NTK with depth - Cora, Citeseer, WebKB}
\begin{figure}[h]
    \centering
    \includegraphics[width=\linewidth]{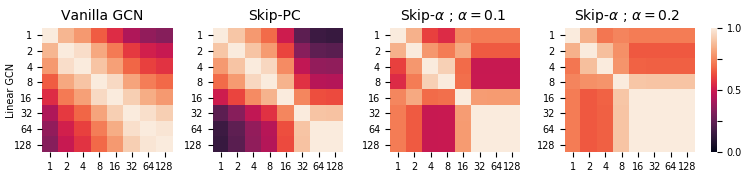}
    \caption{Convergence of NTK with depth for all the discussed linear architectures evaluated on Cora dataset.}
    \label{fig:conv_lin_cora}
\end{figure}

\begin{figure}[h]
    \centering
    \includegraphics[width=\linewidth]{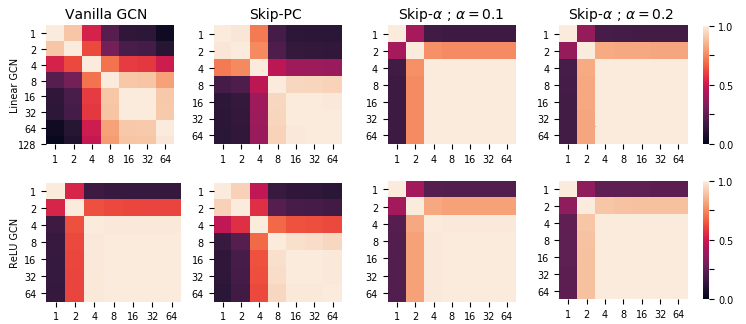}
    \caption{Convergence of NTK with depth for all the discussed linear and ReLU architectures evaluated on Citeseer dataset.}
    \label{fig:conv_citeseer}
\end{figure}

\begin{figure}[h]
    \centering
    \includegraphics[width=\linewidth]{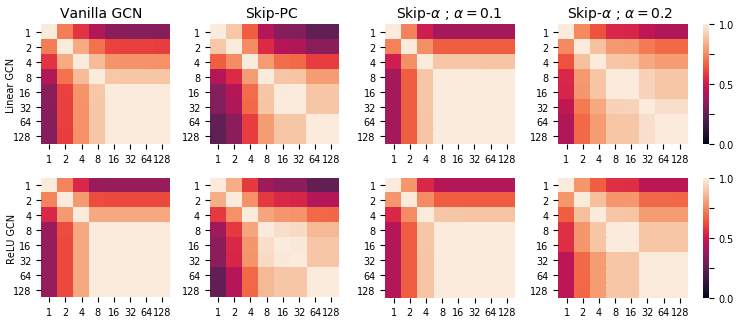}
    \caption{Convergence of NTK with depth for all the discussed linear and ReLU architectures evaluated on WebKB dataset.}
    \label{fig:conv_webkb}
\end{figure}

We presented the convergence of NTK with depth for ReLU GCN with and without skip connections evaluated on Cora dataset in Figure~\ref{fig:ntk_heatmap}.
Here, we present the convergence plot for Linear GCN evaluated on Cora and all discussed linear and ReLU networks evaluated Citeseer and WebKB.
The observation is similar to the discussion in Section~\ref{sec:convergence}.
Figures~\ref{fig:conv_lin_cora}, \ref{fig:conv_citeseer} and \ref{fig:conv_webkb} show the convergence plots for linear GCN evaluated on Cora, ReLU and linear GCNs with and without skip connections for Citeseer and WebKB, respectively.

\end{document}